\documentclass[10pt,twocolumn,letterpaper]{article}

\usepackage{cvpr}              

\usepackage[dvipsnames]{xcolor}

\definecolor{cvprblue}{rgb}{0.21,0.49,0.74}
\usepackage[pagebackref,breaklinks,colorlinks,citecolor=cvprblue]{hyperref}

\usepackage{booktabs} 
\usepackage{colortbl}  %
\usepackage{xcolor}

\usepackage[accsupp]{axessibility}

\def\paperID{2631} 
\def\confName{CVPR}
\def\confYear{2024}

\title{Open-Vocabulary Video Anomaly Detection}

\author{
    Peng Wu\textsuperscript{\rm 1}, Xuerong Zhou\textsuperscript{\rm 1}, Guansong Pang\textsuperscript{\rm 2}\thanks{Corresponding Authors}, Yujia Sun\textsuperscript{\rm 3}, Jing Liu\textsuperscript{\rm 3}, Peng Wang\textsuperscript{\rm 1}$^\ast$, Yanning Zhang\textsuperscript{\rm 1}\\
    \textsuperscript{\rm 1}Northwestern Polytechnical University, 
    \textsuperscript{\rm 2}Singapore Management University, 
    \textsuperscript{\rm 3}Xidian University\\
    {\tt\small \{xdwupeng, zxr2333\}@gmail.com, gspang@smu.edu.sg, yjsun@stu.xidian.edu.cn}\\
    {\tt\small neouma@163.com, \{peng.wang, ynzhang\}@nwpu.edu.cn}
    }

\begin{document}
\maketitle
\begin{abstract}
Current video anomaly detection (VAD) approaches with weak supervisions are inherently limited to a closed-set setting and may struggle in open-world applications where there can be anomaly categories in the test data unseen during training. A few recent studies attempt to tackle a more realistic setting, open-set VAD, which aims to detect unseen anomalies given seen anomalies and normal videos. However, such a setting focuses on predicting frame anomaly scores, having no ability to recognize the specific categories of anomalies, despite the fact that this ability is essential for building more informed video surveillance systems. This paper takes a step further and explores open-vocabulary video anomaly detection (OVVAD), in which we aim to leverage pre-trained large models to detect and categorize seen and unseen anomalies. To this end, we propose a model that decouples OVVAD into two mutually complementary tasks -- class-agnostic detection and class-specific classification -- and jointly optimizes both tasks. Particularly, we devise a semantic knowledge injection module to introduce semantic knowledge from large language models for the detection task, and design a novel anomaly synthesis module to generate pseudo unseen anomaly videos with the help of large vision generation models for the classification task. These semantic knowledge and synthesis anomalies substantially extend our model's capability in detecting and categorizing a variety of seen and unseen anomalies.   
Extensive experiments on three widely-used benchmarks 
demonstrate our model achieves state-of-the-art performance on OVVAD task.
\end{abstract}
\section{Introduction}\label{sec1}
Video anomaly detection (VAD), which aims at detecting unusual events that do not conform to expected patterns, has become a growing concern in both academia and industry communities due to its promising application prospects in, such as, intelligent video surveillance and video content review. Through several years of vigorous development, VAD has made significant progress with many works continuously emerging.

\begin{figure}[t]%
\centering
\includegraphics[width=0.99\linewidth]{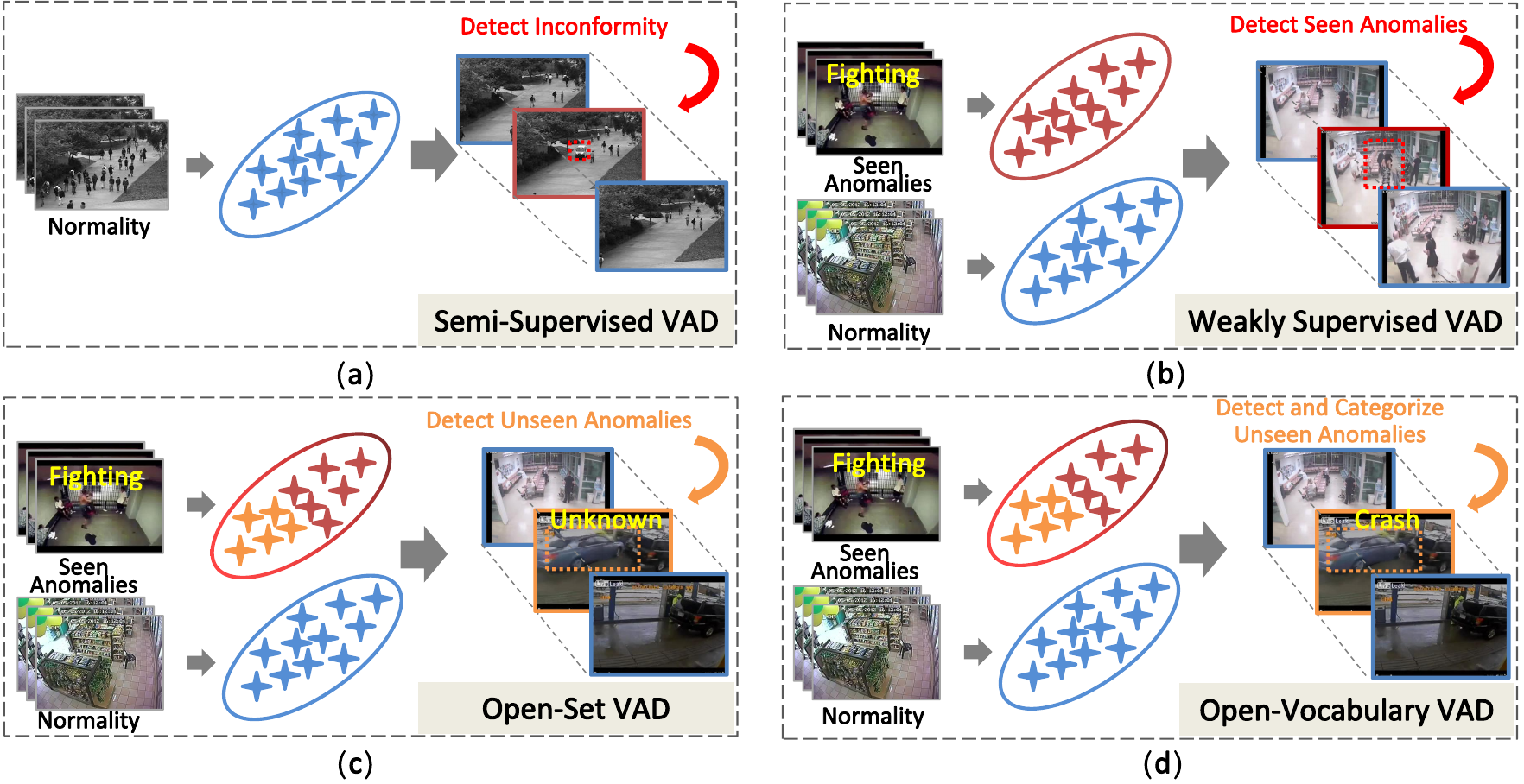}
\caption{Comparison of different VAD tasks.}\label{comparison}
\vspace{-0.3cm} 
\end{figure}

Traditional VAD can be broadly classified into two types based on the supervised mode, i.e., semi-supervised VAD~\cite{liu2018future} and weakly supervised VAD~\cite{sultani2018real}. The main difference between them lies in the availability of abnormal training samples. Although they are different in terms of supervised mode and model design, both can be roughly regarded as classification tasks. In the case of semi-supervised VAD, it falls under the category of one-class classification, while weakly supervised VAD pertains to binary classification. Specifically, semi-supervised VAD assumes that only normal samples are available during the training stage, and the test samples which do not conform to these normal training samples are identified as anomalies, as shown in~\cref{comparison}(a). Most existing methods essentially endeavor to learn the one-class pattern, i.e., normal pattern, by means of one-class classifiers~\cite{wu2019deep} or self-supervised learning technique, e.g. frame reconstruction~\cite{hasan2016learning}, frame prediction~\cite{liu2018future}, jigsaw puzzles~\cite{wang2022video}, etc. Similarly, as illustrated in~\cref{comparison}(b), weakly supervised VAD can be seen as a binary classification task with the assumption that both normal and abnormal samples are available during the training phase but the precise temporal annotation of abnormal events are unknown. Previous approaches widely adopt a binary classifier with the multiple instance learning (MIL)~\cite{sultani2018real} or Top-K mechanism~\cite{pu2023learning} to discriminate between normal and abnormal events. In general, existing approaches of both semi-supervised and weakly supervision VAD
restricts their focus to classification and use corresponding discriminator to categorize each video frame. While these practices have achieved significant success on several widely-used benchmarks, they are limited to detecting a closed set of anomaly categories and are unable to handle arbitrary unseen anomalies. This limitation restricts their application in open-world scenarios and poses a risk of increasing missing reports, as many real-world anomalies in actual deployment are not present in the training data.

To address this issue, a few recent works explore a whole new line of VAD, i.e., open-set VAD~\cite{zhu2022towards, ding2022catching, acsintoae2022ubnormal, zhu2023anomaly}. The core purpose of open-set VAD is to train a model with normal and seen abnormal samples to detect unseen anomalies (see~\cref{comparison}(c)). For example, abnormal training samples only includes fighting and shooting events, and it is expected that the trained model can detect abnormal events that occur in the road accident scene. Compared to traditional VAD, open-set VAD breaks out of the close-set dilemma and then possesses ability to deal with open-world problems. Although these works partly reveal their open-world capacity, 
they fall short in addressing semantic understanding of the abnormal category, which leads to the ambiguous detection process in the open world.

Recently, large language/vision model pre-training~\cite{radford2021learning, jia2021scaling, rombach2022high, zhou2023anomalyclip} has been phenomenally successful across a wide range of downstream tasks~\cite{wu2023towards,zareian2021open, kim2023region, zhou2023zegclip, qin2023freeseg, ni2022expanding, ju2022prompting, nag2022zero, weng2023transforming, ju2023multi} on account of its learned cross-modal prior knowledge and powerful transfer learning ability, which also allow us to tackle open-vocabulary video anomaly detection (OVVAD). Therefore, in this paper, we propose a novel model built upon large pre-trained vision/language models for OVVAD that aims to detect and categorize seen and unseen anomalies, as shown in~\cref{comparison}(d). Compared to previous VAD, OVVAD has high value to applications as it can provide more informed, fine-grained detection results, but it is more challenging since that 1) it not only needs to detect but also categorize the anomalies; 2) it needs to tackle seen (base) as well as unseen (novel) anomalies. To address these challenges, we explicitly disentangle the OVVAD task into two mutually complementary sub-tasks: one is class-agnostic detection, while another one is class-specific categorization. To improve the class-agnostic detection, we make efforts from two aspects. We first introduce a nearly weight-free temporal adapter (TA) module to model temporal relationships, and then introduce a novel semantic knowledge injection (SKI) module designed to incorporate textual knowledge into visual signals with assistance of large language models. To enhance the class-specific categorization, we take inspirations from the contrastive language-image pre-training (CLIP) model~\cite{radford2021learning}, and use a scalable way to categorize anomalies, i.e., the alignment between textual labels and videos, and furthermore we design a novel anomaly synthesis (NAS) module to generate vision (e.g., images and videos) materials to assist the model better identify novel anomalies. Based on these operations, our model achieves state-of-the-art performance on three popular benchmarks for OVVAD, 
attaining 86.40\% AUC, 66.53\% AP and 62.94\% AUC on UCF-Crime~\cite{sultani2018real}, XD-Violence~\cite{wu2020not} and UBnormal~\cite{acsintoae2022ubnormal}, respectively.

We summarize our contributions as follows:
\begin{itemize}
    \item We explore video anomaly detection under a challenging yet practically important open-vocabulary setting. To our knowledge, this is the first work for OVVAD.
    \item We then propose a model built on top of pre-trained large models that disentangles the OVVAD task into two mutually complementary sub-tasks -- class-agnostic detection and class-specific categorization -- and jointly optimizes them for accurate OVVAD.
    \item In the class-agnostic detection task, we design a nearly weight-free temporal adapter module and a semantic knowledge injection module for substantially-enhanced normal/abnormal frame detection.
    \item In the fine-grained anomaly classification task, we introduce a novel anomaly synthesis module to generate pseudo unseen anomaly videos for accurate classification of novel anomaly types.
\end{itemize}

\section{Related Work}\label{sec2}
\noindent\textbf{Semi-supervised VAD.}  
Mainstream solutions are to build a normal pattern by self-supervised manner (e.g., reconstruction and prediction) or one-class manner. As for the self-supervised manner~\cite{yu2020cloze, georgescu2021anomaly, yan2023towards}, reconstruction-based approaches~\cite{cong2011sparse, lu2013abnormal, zhao2017spatio, luo2017remembering, ristea2023self, sun2023hierarchical, yang2023video} typically leverage encoder-decoder frameworks to reconstruct normal events and compute the reconstruction errors, and these events with large reconstruction error are classified as anomalies. 
Follow-up prediction-based approaches~\cite{liu2018future, liu2021hybrid} focuses on predicting the future frame with previous video frames and determine whether it is an anomaly frame by calculating the difference between the predicted frame and the actual frame. Recent work~\cite{shi2023video} combined reconstruction- and prediction-based approaches to improve detection performance. As for one-class models, some works endeavors to learn normal patterns by making use of one-class frameworks~\cite{sabokrou2018adversarially}, e.g., one-class support vector machine and its extension (OCSVM~\cite{scholkopf1999support}, SVDD~\cite{wu2019deep}, GODS~\cite{wang2019gods}). 

\noindent\textbf{Weakly supervised VAD.} 
In contrast to semi-supervised VAD, weakly supervised VAD~\cite{huang2022weakly, sun2023long} consists of normal as well as abnormal samples, which can be regarded as a binary classification task and aims to detect anomalies at frame level under the limitation of temporal annotations. As a pioneer work, Sultani et al.~\cite{sultani2018real} first proposed a large-scale benchmark and trained a lightweight network with MIL mechanism. Then Zhong et al.~\cite{zhong2019graph} proposed a graph convolutional network based approach to capture the similarity relations and temporal relations across frames. Tian et al.~\cite{tian2021weakly} introduced self-attention blocks and pyramid dilated convolution layers to capture multi-scale temporal relations. Wu et al.~\cite{wu2020not, wu2022weakly} built the largest-scale benchmark that includes audio-visual signals and proposed a multi-task model to deal with coarse- and fine-grained VAD. Zaheer et al.~\cite{zaheer2020claws} presented a clustering assisted weakly supervised framework with novel normalcy suppression mechanism. Li et al.~\cite{li2022self} proposed a transformer-based network with self-training multi-sequence learning. Zhang et al.~\cite{zhang2023exploiting} attempted to exploit the completeness and uncertainty of pseudo labels. 
The above approaches simply used video or audio inputs encoded by pre-trained models, such as C3D~\cite{tran2015learning} and I3D~\cite{carreira2017quo}, although a few works~\cite{joo2023clip, lv2023unbiased, wu2023vadclip} introduced CLIP models to weakly-supervised VAD task, they simply used its powerful visual features and ignored the zero-shot ability of CLIP. 

\noindent\textbf{Open-set VAD.} 
VAD task naturally exists an open-world requirement. 
Faced with an open-world requirement, traditional semi-supervised works are more prone to producing large false alarms, while weak-supervised works are effective in detecting known anomalies but could fail in unseen anomalies. Open-set VAD aims to train the model based on normality and seen anomalies, and attempts to detect unseen anomalies. Acsintoae et al.~\cite{acsintoae2022ubnormal} developed the first benchmark called UBnormal for supervised open-set VAD task. Zhu et al.~\cite{zhu2022towards} proposed an approach to deal with open-set VAD task by integrating evidential deep learning and normalizing flows into a MIL framework. Besides, Ding et al.~\cite{ding2022catching} proposed a multi-head network based model to learn the disentangled anomaly representations, with each head dedicated to capturing one specific type of anomaly. Compared to our model, these above works mainly devote themselves to open-world detection and overlook anomaly categorization, moreover, these works also fail to take full advantage of pre-trained models.%


\begin{figure*}[h]%
\centering
\includegraphics[width=0.75\textwidth]{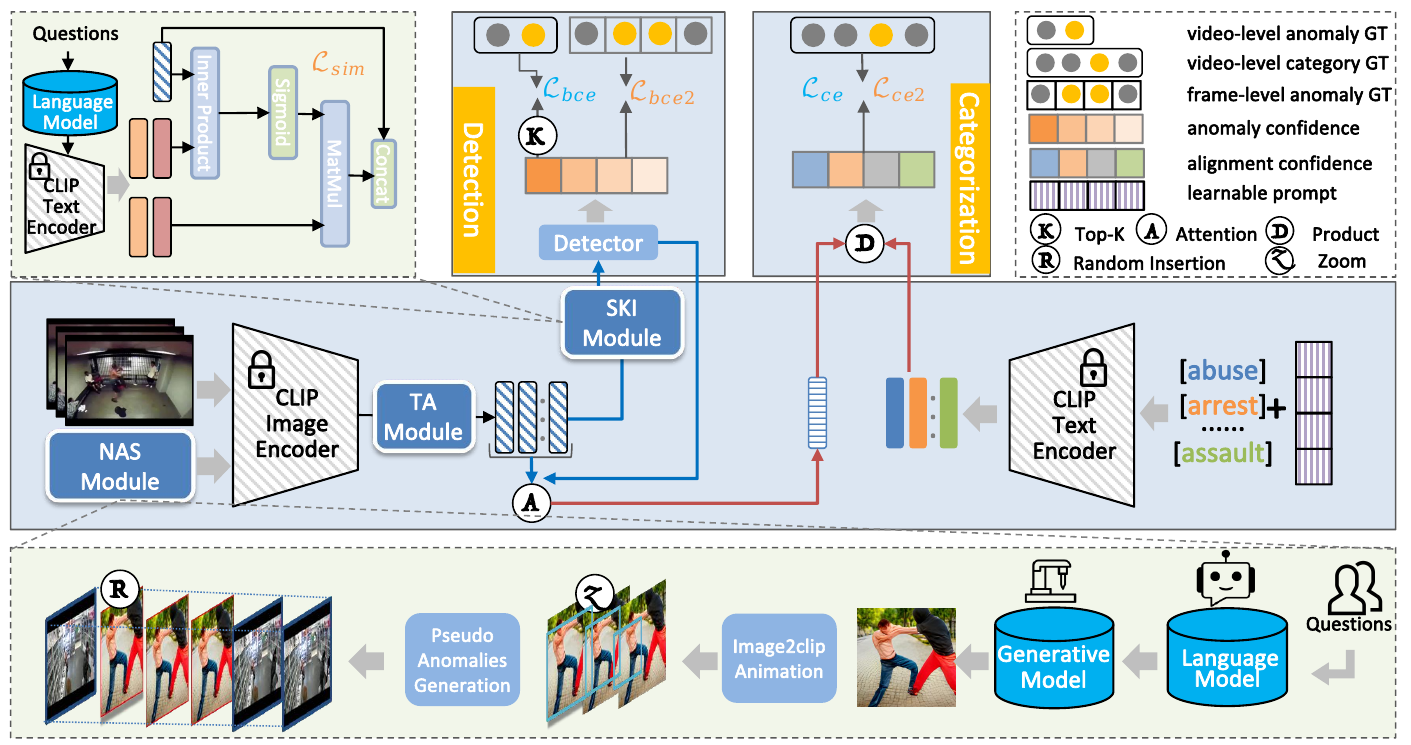}
\caption{Overview of our proposed framework.}\label{method}
\end{figure*}

\section{Method}\label{sec3}

\textbf{Problem Statement.} The studied problem, OVVAD, can be formally stated as follows. Suppose we are given a set of training samples $\mathcal{X}=\{x_i\}_{i=1}^{N+A}$, where $\mathcal{X}_n=\{x_i\}_i^{N}$ is the set of normal samples and $\mathcal{X}_a=\{x_i\}_{i=N+1}^{N+A}$ is the set of abnormal samples. For each sample $x_i$ in $\mathcal{X}_a$, it has a corresponding video-level category label $y_i$, $y_i \in C_{base}$, Here, $C_{base}$ represents the set of base (seen) anomaly categories, and $\mathcal{C}$ is the union of $C_{base}$ and $C_{novel}$, where $C_{novel}$ stands for the set of novel (unseen) anomaly categories.
Based on the training samples $\mathcal{X}$, the objective is to train a model capable of detecting and categorizing both base and novel anomalies. Specifically, the goal of model is to predict anomaly confidence for each frame, and identify the anomaly category if anomalies are present in the video.

\subsection{Overall Framework}\label{subsec32}
Traditional methods based on close-set classifications are less likely to deal with VAD under the open-vocabulary scenario. To this end, we leverage language-image pre-training models, e.g., CLIP, as the foundation thanks to its powerful zero-shot generalization ability. As illustrated in~\cref{method}, given a training video, we first feed it into image encoder of CLIP $\Phi_{CLIP-v}$ to obtain frame-level features $x_f$ with shape of $n \times c$, where $n$ is the number of video frames, and $c$ is the feature dimension. Then these features pass through TA module, SKI module and a detector to produce frame-level anomaly confidence $p$, this pipeline is mainly applied to class-agnostic detection task. On the other hand, for class-specific categorization, we take inspirations from other open-vocabulary works across different vision tasks~\cite{wang2021actionclip, rao2022denseclip, zhou2022learning} and use cross-modal alignment mechanism. Specifically, we first generate a video-level aggregated feature across frame-level features, then also generate textual features/embeddings of anomaly categories, finally, we estimate the anomaly category based on alignments between video-level features and textual features. Moreover, we introduce NAS module to generate potential novel anomalies with the assistance of large language models (LLM) and AI-generated content models (AIGC) for novel category identification achievement.

\subsection{Temporal Adapter Module}\label{subsec33}
Temporal dependencies plays a vital role in VAD~\cite{zhou2023dual, wu2021learning}. In this work, we employ the frozen image encoder of CLIP to attain vision features, but it lacks consideration of temporal dependencies since CLIP is pre-trained on image-text pairs. To bridge the gap between images and videos, the use of a temporal transformer~\cite{ju2022prompting, ni2022expanding} has emerged as a routine practice in recent studies. However, such a paradigm suffers from a clear performance degradation on novel categories~\cite{ju2022prompting, rasheed2023fine}, the possible reason is that additional parameters in temporal transformer could specialise on the training set, thus harming the generalisation towards novel categories. Therefore, we design a nearly weight-free temporal adapter for temporal dependencies, which is built on top of classical graph convolutional networks. Mathematically, it can be presented as follows,
\begin{equation}
  x_t = LN(softmax\left(H\right)x_f)
\end{equation}
where $LN$ is the layer normalization operation, $H$ is the adjacency matrix, the softmax normalization is used to ensure the sum of each row of $H$ equals to one. Such a design is used to capture contextual dependencies based on positional distance between each two frames. The adjacency matrix is calculated as follows:
\begin{equation}
H_{(i,j)}=\frac{-|i-j|}{\sigma }\label{con:position}
\end{equation}
the proximity relation between $i^{th}$ and $j^{th}$ frames only determined by their relative temporal position. $\sigma$ is a hyper-parameter to control the range of influence of distance relation. According to this formula, the closer the temporal distance between two frames, the higher proximity relation the score, otherwise the lower. Notably, across TA module, only layer normalization involves few parameters.

\subsection{Semantic Knowledge Injection Module}\label{subsec34}
Human often make use of prior knowledge when perceiving the environment, for example, we can infer the presence of a fire based on the smell and smoke without directly seeing the flames. Building on this idea, we propose SKI module to explicitly introduce additional semantic knowledge for assisting visual detection. As depicted in~\cref{method}, for normal events in videos, we prompt the large-scale language models, e.g., ChatGPT~\cite{brown2020language} and SparkDesk~\footnote{https://xinghuo.xfyun.cn}, with a fixed template, to obtain about common scenarios and actions, such as, street, park, shopping hall, walking, running, working, etc. Likewise, we generate additional words related to anomaly scenes, including terms like explosion, burst, firelight, etc. Finally, we obtain several phrase lists denoted by $M_{prior}$, which consists of noun words (scenes) and verb words (actions). With $M_{prior}$ in hands, we exploit the text encoder of CLIP to extract textual embeddings as the semantic knowledge, which is show as follows,
\begin{equation}
F_{text} = \Phi_{CLIP-t}\left(\Phi_{token} (M_{prior})\right)
\label{con:token1}
\end{equation}
where $F_{text}\in \mathcal{R}^{l\times c}$, $\Phi_{CLIP-t}$ denotes the text encoder of CLIP, and $\Phi_{token}$ refers to the language tokenizer that converts words into vectors.

Then, towards the goal of effectively incorporating these semantic knowledge into visual information to boost anomaly detection, we design a cross-modal injection strategy. This strategy encourage visual signals to seek related semantic knowledge and integrate it into the process. Such an operation is demonstrated as,
\begin{equation}
F_{know} = {sigmoid\left(x_tF_{text}^{\top}\right)F_{text}}/{l}
\label{con:ski}
\end{equation}
where $F_{know}\in\mathcal{R}^{n\times c}$, and we employ sigmoid instead of softmax to ensure that visual signals can encompass more relevant semantic concepts.

Finally, we concatenate $F_{know}$ and $x_{t}$ creating an input that contains both visual information and integrated semantic knowledge. We feed this input into a binary detector to generate anomaly confidence for class-agnostic detection.

\subsection{Novel Anomaly Synthesis Module}\label{subsec35}
While current pre-trained vision-language models, such as, CLIP, possess impressive zero-shot capacities, their zero-shot performance on various downstream tasks, especially video-related ones, remains far from satisfactory. For the same reason, our model, which is built on these pre-trained vision-language models, is trained on base anomalies and normal samples, making it susceptible to a generalization deficiency when faced with novel anomalies. With the advent of large generative models, generating samples as pseudo training data has become a feasible solution~\cite{ni2022imaginarynet, liu2023generating}. Consequently, we propose NAS module to generate a series of pseudo novel anomalies based solely on potential anomaly categories. We then leverage these samples to fine-tune the proposed model for improved categorization and detection of novel anomalies. 
On the whole, NAS module consists of three key processes: 

1) Initially, we prompt LLMs (e.g., ChatGPT, ERNIE Bot~\cite{sun2020ernie}) with pre-defined templates $prompt_{gen}$ like \textit{generate ten shorter scene descriptions about ``Fighting'' in real world} to produce textual descriptions of potential novel categories. We then employ AIGC models, e.g., DALL·E mini~\cite{ramesh2022hierarchical}, Gen-2~\cite{esser2023structure}, to generate corresponding images or short videos. This can be represented as follows,
\begin{equation}
V_{gen} = \Phi_{AIGC}\left(\Phi_{LLM} (prompt_{gen})\right)
\label{con:nas}
\end{equation}
where $V_{gen}$ is the combination of generated images ($I_{gen}$) and short videos ($S_{gen}$).

2) Subsequently, for $I_{gen}$, we draw inspiration from~\cite{liu2022animating} and introduce a simple yet effective animation strategy to convert single images into video clips that simulates scene changes. Specifically, given an image, we employ the center crop mechanism with different crop ratios to select corresponding image regions, then resize these regions back to original size and cascade them to create new video clips $S_{cat}$.

3) Finally, to mimic real-world situation where anomaly videos are generally long and untrimmed, we introduce the third step, pseudo anomaly synthesis, by inserting $S_{cat}$ or $S_{gen}$ into randomly selected normal videos. Moreover, the insertion position is also randomly chosen. This process yields the final pseudo anomaly samples $\mathcal{V}_{nas}$. Refer to supplementary materials for detailed descriptions and results.

With $\mathcal{V}_{nas}$ in hands, we fine-tune our model, which was initially trained on $\mathcal{X}$, to enhance its generalization capacities for novel anomalies.

\subsection{Objective Functions}\label{subsec36}

\subsubsection{Training stage without pseudo anomaly samples}
For class-agnostic detection, following previous VAD works~\cite{wu2021learning, pu2023learning}, we use the Top-K mechanism to select the top $K$ high anomaly confidences in both abnormal and normal videos. We compute the average values of these selections and feed the average values into the sigmoid function as the video-level predictions. Here, we set $K=n/16$ for abnormal videos and $K=n$ for normal videos. Finally, we compute binary cross entropy $L_{bce}$ between video-level prediction and binary labels. 

In regard to class-specific categorization, we compute the similarity between aggregated video-level features and textual category embeddings to derive video-level classification predictions. 
We also use a cross entropy loss function to compute the video-level categorization loss $L_{ce}$. 
Given that OVVAD is a weakly supervised task, we can not obtain video-level aggregated features directly from frame-level annotations. Following~\cite{wu2021learning}, we employ a soft attention based aggregation, as shown below,
\begin{equation}
 \begin{split}
  x_{agg}=softmax(p)^{\top}x_{t}
  \label{con:att}
 \end{split}
\end{equation}
For textual category embeddings, we are inspired by CoOp\cite{zhou2022learning} and append the learnable prompt to original category embeddings. 

For the parameters of SKI module, namely $F_{text}$, we aim for explicit optimization during the training stage. We intend to distinguish between normal knowledge embeddings and abnormal knowledge embeddings. For normal videos, we expect their visual features have higher similarities with normal knowledge embeddings and lower similarities with abnormal knowledge embeddings. 
To this end, we first extract the similarity matrix between each video and textual knowledge embeddings, and then select the top $10\%$ highest scores for each frame and compute the average value, finally, we apply the cross-entropy-base loss $L_{sim-n}$. For abnormal videos, we anticipate the high similarities between abnormal knowledge embeddings and abnormal video-frame features. Since precise frame-level annotations are absent under weak supervision, we employ a hard attention based selection mechanism know as Top-K to locate abnormal regions. The same operations are then performed to compute the loss $L_{sim-a}$.



Overall, during the training phase, we employ three loss functions, with the total loss function given as:
\begin{equation}
 \begin{split}
  L_{train} = L_{bce} + L_{ce} + L_{sim}
  \label{con:total1}
 \end{split}
\end{equation}
where $L_{sim}$ is the sum of $L_{sim-n}$ and $L_{sim-a}$.

\subsubsection{Fine-tuning stage with pseudo anomaly samples}
After obtaining $\mathcal{V}_{nas}$ from NAS module, we proceed with fine-tuning our model. $\mathcal{V}_{nas}$ is synthetic, providing us with frame-level annotations and allowing us to optimize our model with full supervisions for detection. For categorization, $L_{ce2}$ remains the same as $L_{ce}$, with the key difference being that labels are available not only for base categories but also for potential novel categories. For detection, $L_{bce2}$ is the binary cross entropy loss at the frame level.

Finally, the total loss function during the fine-tuning phase is shown as:
\begin{equation}
 \begin{split}
  L_{tune} = L_{bce2} + L_{ce2} + \lambda(L_{bce}+L_{ce})
  \label{con:total1}
 \end{split}
\end{equation}

\section{Experiments}\label{sec4}

\subsection{Experiment Setup}\label{subsec41}
\noindent\textbf{Datasets.} \textbf{UCF-Crime}~\cite{sultani2018real} is a large-scale VAD dataset for surveillance scenes, containing 13 types of abnormal events. 800 normal videos and 810 abnormal videos are provided for training, and the remaining 140 normal videos and 150 abnormal videos for test. \textbf{XD-Violence}~\cite{wu2020not} is the largest VAD benchmark to date, it contains 6 anomalous categories with 3954 videos for training and 800 videos for test. To align with our model, which supports single-category identification, we exclude videos with multiple categories in XD-Violence. 
\textbf{UBnormal}~\cite{acsintoae2022ubnormal} is a synthesized benchmark 
which defines seven types of normal events and 22 types of abnormal events. During training, only 7 abnormal categories are visible, while 12 abnormal categories are used for test. 

\noindent\textbf{Evaluation metrics.} OVVAD entails detecting and categorizing anomalies. To assess detection performance, we employ standard metrics from previous works~\cite{sultani2018real, wu2020not}. For UCF-Crime and UBnormal, we use the area under the curve of the frame-level receiver operating characteristic (AUC) to evaluate performance. For XD-Violence, we utilize AUC of the frame-level precision-recall curve (AP). For classification, we report the video-level TOP1 accuracy for abnormal test videos on both UCF-Crime and XD-Violence. UBnormal lacks category labels, so we exclusively report AUC results.
During the test phase, we provide these metrics for the entire set of categories, as well as separately for base and novel categories, on both UCF-Crime and XD-Violence.

\noindent\textbf{Implementation Details.} 
The proposed model is implemented using PyTorch and trained on single RTX3090 GPU. 
The frozen image encoder and text encoder stem from pre-trained CLIP(ViT-B/16)~\cite{dosovitskiy2020image} model. The detector is a modified feed-forward network (FFN) layer in Transformer with ReLU replaced by GeLU.
In line with existing works, we process 1 out of 16 frames for each video, 
and during the training phase, the maximum video length is set to 256. 
For model optimization, we use AdamW optimizer to train the model with learning rate of $1e^{-4}$ and train epoch of 20. The batch size is set to 64, consisting of an equal number of normal and abnormal samples. During the fine-tune phase with pseudo novel anomalies, the learning rate is set to $1e^{-5}$ on UBnormal and $5e^{-6}$ on UCF-Crime and XD-Violence. The fine-tuning process spans 10 epochs, with each batch containing 10 pseudo novel anomaly videos and 10 base anomaly videos. 
$\sigma$ is set to 0.07 across all situations, and $\lambda$ is set as $1e^{-1}$ on UCF-Crime, $1e^{0}$ on XD-Violence and UBnormal, respectively. 

\begin{table}[t]
  \centering
  \resizebox{\linewidth}{!}{
  \begin{tabular}{l|lccc}
    \toprule
    Mode&Method&AUC(\%)&AUC$_b$(\%)&AUC$_n$(\%)\\
    \midrule
     &SVM baseline &50.10 & N/A&N/A\\
     Semi&OCSVM\cite{scholkopf1999support} &63.20 & N/A&N/A\\
     &Hasan et al.\cite{hasan2016learning} & 51.20& N/A&N/A\\
    \midrule
    &Sultani et al.$^\dagger$\cite{sultani2018real}& 84.14& N/A&N/A\\
    &Wu et al.$^\dagger$\cite{wu2020not} &84.57 & N/A&N/A\\
    &AVVD$^\dagger$\cite{wu2022weakly} & 82.45&N/A& N/A\\
    Weak&RTFM$^\dagger$\cite{tian2021weakly} &85.66 & N/A&N/A\\
    &DMU$^\dagger$\cite{zhou2023dual} & 86.75 &N/A&N/A\\
    &UMIL$^\dagger$\cite{lv2023unbiased} & 86.75& N/A&N/A\\
    &CLIP-TSA$^\dagger$\cite{joo2023clip} & 87.58& N/A &N/A\\
    \midrule
    \midrule
    &Zhu et al.$^\ast$\cite{zhu2022towards} & 78.82& N/A&N/A\\
    &Sultani et al.\cite{sultani2018real}& 78.25& 86.31&80.12\\
    Weak&Wu et al.\cite{wu2020not} &82.24 & 90.62&84.13\\
    &RTFM\cite{tian2021weakly} &84.47 & 92.54&85.87\\
    &DMU\cite{zhou2023dual} & 85.14&93.52&86.24\\
    &\cellcolor{gray!20}\textbf{Ours} & \cellcolor{gray!20}\textbf{86.40} &\cellcolor{gray!20}\textbf{93.80}&\cellcolor{gray!20}\textbf{88.20} \\
  \bottomrule
\end{tabular}}
\caption{AUC Comparisons on UCF-Crime.}
  \label{tab:ucf}
\vspace{-0.3cm} 
\end{table}

\begin{table}[!t]
  \centering
  \scalebox{0.9}{
  \begin{tabular}{l|lccc}
    \toprule
    Mode&Method&AP(\%)&AP$_b$(\%)&AP$_n$(\%)\\
    \midrule
     &SVM baseline &50.80& N/A&N/A\\
     Semi&OCSVM\cite{scholkopf1999support} &28.63& N/A&N/A\\
     &Hasan et al.\cite{hasan2016learning} & 31.25& N/A&N/A\\
    \midrule
    &Sultani et al.$^\dagger$\cite{sultani2018real}& 75.18& N/A&N/A\\ 
    &Wu et al.$^\dagger$\cite{wu2020not} &80.00& N/A&N/A\\
    Weak&RTFM$^\dagger$\cite{tian2021weakly} &78.27& N/A&N/A\\
    &AVVD$^\dagger$\cite{wu2022weakly}&78.10& N/A&N/A\\
    &DMU$^\dagger$\cite{zhou2023dual} & 82.41& N/A&N/A\\
    &CLIP-TSA$^\dagger$\cite{joo2023clip} & 82.17& N/A&N/A\\
    \midrule
    \midrule
    &Zhu et al.$^\ast$\cite{zhu2022towards} & 64.40& N/A&N/A\\
    &Sultani et al.\cite{sultani2018real}& 52.26&51.25&54.64 \\
    Weak&Wu et al.\cite{wu2020not} &55.43 &52.94&64.10 \\
    &RTFM\cite{tian2021weakly} &58.99 & 55.72&65.97\\ 
    &DMU\cite{zhou2023dual} & 63.90& \textbf{60.12}&71.63\\
    &\cellcolor{gray!20}\textbf{Ours} & \cellcolor{gray!20}\textbf{66.53} & \cellcolor{gray!20}57.10& \cellcolor{gray!20}\textbf{76.03} \\
  \bottomrule
\end{tabular}}
\caption{AP Comparisons on XD-Violence.}
  \label{tab:xd}
\vspace{-0.3cm} 
\end{table}

\begin{table}[!t]
  \centering
  \scalebox{0.9}{
  \begin{tabular}{l|lc}
    \toprule
    Mode&Method&AUC(\%)\\
    \midrule
    Semi&Georgescu et al.\cite{georgescu2021anomaly}&59.30\\
    \midrule
    &Georgescu et al.\cite{georgescu2021anomaly}+anomalies &61.30\\
    &Sultani et al.\cite{sultani2018real}& 50.30\\
    Weak&Wu et al.\cite{wu2020not} &53.70\\ 
    &RTFM\cite{tian2021weakly} &60.94\\ 
    &DMU\cite{zhou2023dual} & 59.91\\
    &\cellcolor{gray!20}\textbf{Ours} & \cellcolor{gray!20}\textbf{62.94} \\
  \bottomrule
\end{tabular}}
  \caption{AUC Comparisons on UBnormal.}
  \label{tab:ub}
\vspace{-0.3cm} 
\end{table}

\subsection{Comparison with State-of-the-Arts}\label{subsec43}
In~\cref{tab:ucf} to~\cref{tab:ub}, we report comparison results with existing approaches on three public benchmarks. Since prior approaches  are designed for close-set VAD, our focus is primarily on the comparison results for open-set detection. For the sake of fairness, most of comparison approaches are re-implemented with the same visual feature as our approach. The symbol $\dagger$ indicates that these approaches follow traditional VAD works and use the entire training set, which includes novel anomaly samples. Consequently, the performance of these approaches outperforms models trained without novel anomaly samples. This underscores the considerable challenge presented by OVVAD from a detection perspective. Regarding the comparison between our approach and other approaches on OVVAD task, we observe that our approach demonstrates distinct advantages over state-of-the-art counterparts. In fact, our model performs on par with the best competitors that make use of the complete training dataset. For example, our approach surpasses the top-performing model, DMU\cite{zhou2023dual}, by 1.26\% AUC on UCF-Crime, 2.63\% AP on XD-Violence, and 3.03\% AUC on UBnormal. Particularly, when it comes to novel categories, our approach exhibits a clear performance advantage compared to other approaches. Notably, Zhu et al.~\cite{zhu2022towards} is the first work to tackle open-set VAD, where the symbol $\ast$ indicates that its category division setup differs from ours. We report its detection results under settings as identical as possible, with the number of novel categories matching ours. Our model outperforms it by a substantial margin on both UCF-Crime and XD-Violence datasets.

\subsection{Ablation Studies}\label{subsec44}
\subsubsection{Contribution of TA module}
As aforementioned, TA module is devised to capture temporal dependencies, thus enhancing class-agnostic detection abilities. To verify the effectiveness of TA module, we conduct experiments and present ablation results in~\cref{tab:ab-ucf} to~\cref{tab:ab-ub}. 
It can be found that the inclusion of TA module, our model achieves a significant performance improvement across various datasets and metrics. More importantly, unlike previous transformer-like temporal modeling modules~\cite{ni2022expanding, ju2022prompting} used on other open-vocabulary tasks, this nearly weight-free designed module also shows a clear gain for novel anomaly categories, 
e.g., adding TA module results in an improvement of 14.47\% AP on XD-Violence.

\subsubsection{Contribution of SKI module}
In this section, we investigate the contribution of SKI module to class-agnostic detection. As reported in~\cref{tab:ab-ucf} to~\cref{tab:ab-ub}, SKI module boosts detection performance on all datasets regardless of whether TA module is introduced or not. Similar to TA module, SKI can also clearly improve performance for novel anomaly categories. The difference with TA module is that SKI module leverage LLMs to explicitly introduce semantic knowledge into visual signals and knowledge helps better distinguish between normal and abnormal events.

\begin{table}[t]
  \centering
  
  \scalebox{0.88}{
    \begin{tabular}{ccc|ccc}
    \toprule
    TA & SKI&NAS& AUC(\%)&AUC$_b$(\%)&AUC$_n$(\%) \\
    \hline
    $\times$ &$\times$ &$\times$ & 84.79&92.75&86.73 \\
    $\surd$&$\times$ &$\times$ & 85.14&93.22&86.79 \\
    $\times$ & $\surd$&$\times$ & 85.04&92.96&86.89 \\
    $\surd$ &$\surd$ &$\times$ &85.81&\textbf{93.85}&87.62  \\
    \hline
    $\surd$ &$\surd$ &$\surd$ &\textbf{86.40}&93.80&\textbf{88.20}  \\
  
    \bottomrule
    \end{tabular}}%
    \caption{Ablations studies with different designed module on UCF-Crime for detection.}
  \label{tab:ab-ucf}%
  \vspace{-0.3cm}
\end{table}%

\begin{table}[!t]
  \centering
  
  \scalebox{0.88}{
    \begin{tabular}{ccc|ccc}
    \toprule
    TA & SKI&NAS& AP(\%)&AP$_b$(\%)&AP$_n$(\%) \\
    \hline
    $\times$ &$\times$ &$\times$ &53.11&54.84&53.69 \\
    $\surd$&$\times$ &$\times$ & 60.13&	59.38&68.16\\
    $\times$ & $\surd$&$\times$ & 56.62&53.03&63.92\\
    $\surd$ &$\surd$ &$\times$ &65.60&\textbf{61.40}&73.67\\
    \hline
    $\surd$ &$\surd$ &$\surd$ &\textbf{66.53}&57.10&\textbf{76.03}  \\
    \bottomrule
    \end{tabular}}%
  \caption{Ablations studies with different designed module on XD-Violence for detection.}
  \label{tab:ab-xd}%
  \vspace{-0.3cm}
\end{table}%

\begin{table}[!t]
  \centering
  
  \scalebox{0.88}{
    \begin{tabular}{ccc|cc}
    \toprule
    TA &SKI&NAS& AUC(\%)&AP(\%) \\
    \hline
    $\times$ &$\times$ &$\times$ &60.51&65.18 \\
    $\surd$&$\times$ &$\times$ & 61.18&67.36\\
    $\times$ & $\surd$&$\times$ &61.93&66.36\\
    $\surd$ &$\surd$ &$\times$ &61.67&67.43\\
    \hline
    $\surd$ &$\surd$ &$\surd$ &\textbf{62.94}&\textbf{68.07}  \\
    \bottomrule
    \end{tabular}}%
  \caption{Ablations studies with different designed module on UBnormal for detection.}
  \label{tab:ab-ub}%
  \vspace{-0.3cm}
\end{table}%

\begin{table}[!t]
  \centering
  \scalebox{0.88}{
    \begin{tabular}{c|ccc}
    \toprule
    & ACC(\%) & ACC$_b$(\%)&ACC$_n$(\%) \\
    \hline
    w/o NAS  &37.86	&43.14	&34.83  \\
    Finetune\_N  & 39.29	&37.25&	\textbf{40.45}  \\
    Finetune\_N+B(Ours) & \textbf{41.43}  & \textbf{49.02} &37.08\\
    \hline
    w/o NAS  & 59.60	&\textbf{91.98}	&15.18 \\
    Finetune\_N &  62.03	&82.06&	\textbf{34.55}\\
    Finetune\_N+B(Ours) & \textbf{64.68}  & 89.31 &30.90\\
    \bottomrule 		
    \end{tabular}}%
  \caption{Ablations studies on UCF-Crime and XD-Violence for categorization.}
  \label{tab: acc}%
  \vspace{-0.3cm}
\end{table}%

\begin{figure*}[t]%
\centering
\includegraphics[width=0.88\linewidth]{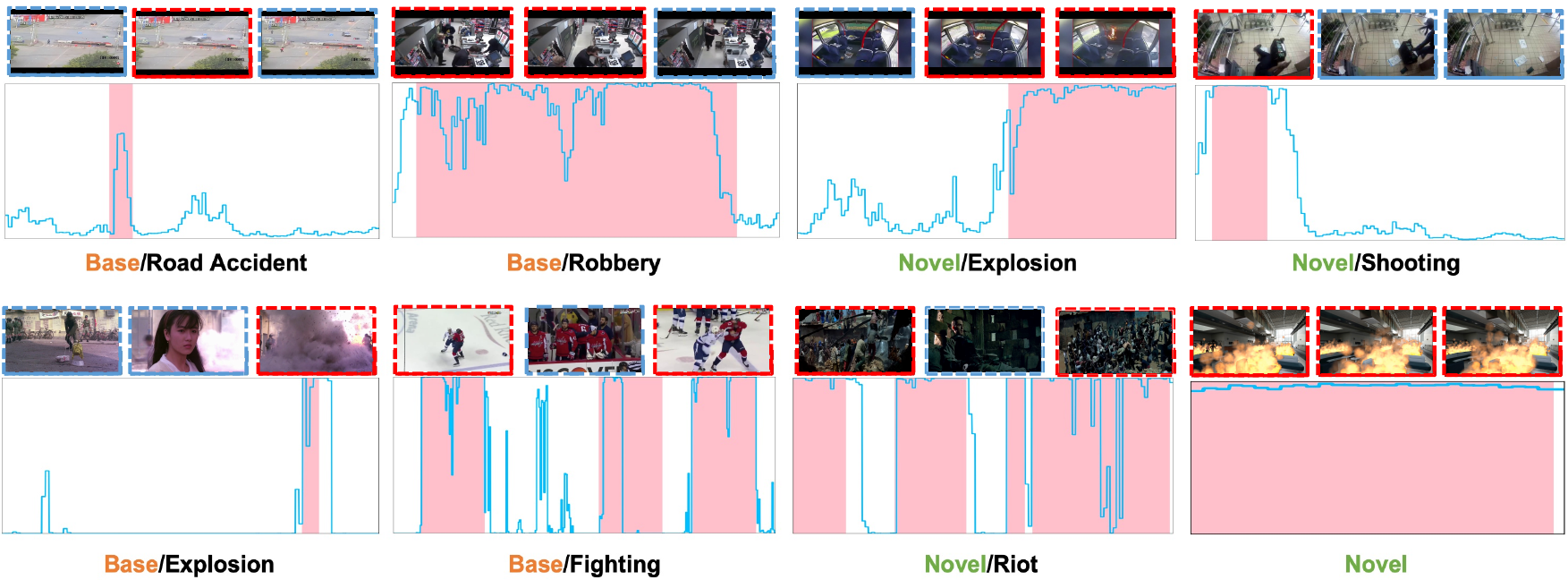}  
\caption{Qualitative results of our model on testing videos. Colored window denotes ground-truth anomalous region.}\label{vis}
\vspace{-0.3cm}
\end{figure*}

\subsubsection{Contribution of NAS module}
From~\cref{tab:ab-ucf} to~\cref{tab:ab-ub}, we can see that, for all test categories and novel anomaly categories, NAS module can obtain a significant performance gain for the class-agnostic detection. 
For base categories, it causes a relatively small performance degradation, which is also observed in other open-vocabulary tasks. We argue that the introduction of pseudo novel samples makes the model pay more attention to these generated novel samples, thus partially diminishing the importance of base categories. Moreover,~\cref{tab: acc} reveals that NAS module also obtains a significant performance gain for the class-specific categorization, especially for novel anomaly categories. Besides, we also found that only using generated novel samples results in a clear performance drop for base anomaly categories during the fine-tuning phase. This illustrates while generated anomaly samples benefit the generalization abilities of our model, it is essential to adopt reasonable and effective fine-tuning schemes.

\begin{table}[!t]
  \centering
  
  \scalebox{0.85}{
    \begin{tabular}{l|cc|cc}
    \toprule
     \qquad ~Test$\Rightarrow$ &\multicolumn{2}{c|}{UCF\_Crime}&\multicolumn{2}{c}{XD\_Violence}\\
     Train$\Downarrow$ & AUC(\%) & ACC(\%)&AP(\%)&ACC(\%) \\
    \hline
    UCF\_Crime  &\cellcolor{gray!20}\textbf{86.05}	&\cellcolor{gray!20}\textbf{45.00}	&63.74&47.90  \\
    XD\_Violence  & 82.42&40.71&\cellcolor{gray!20}\textbf{82.86}&\cellcolor{gray!20}\textbf{88.96}  \\
 
    \bottomrule 		
    \end{tabular}}%
  \caption{Cross-dataset results on UCF-Crime and XD-Violence.}
  \label{tab:cross}%
  \vspace{-0.4cm}
\end{table}%

\subsubsection{Analysis of cross-dataset ability}
To further investigate the zero-shot abilities of our model, we conducted experiments where we train our model under the cross-dataset setup. 
In this case, we take UCF-Crime and XD-Violence as examples. These datasets have some overlapping categories but completely different sources, with UCF-Crime developed from surveillance videos and XD-Violence collected from movies and online videos. From the evaluation results in~\cref{tab:cross}, we can draw the following conclusions: First, our model achieves better performance with the whole training samples. 
Second, the cross-dataset test results show that our model can compete with or outperform current approaches on both UCF-Crime and XD-Violence, further validating the favorable generalization abilities of the proposed model.

\subsection{Qualitative Results}\label{subsec45}

\begin{figure}[t]%
\centering
\includegraphics[width=0.9\linewidth]{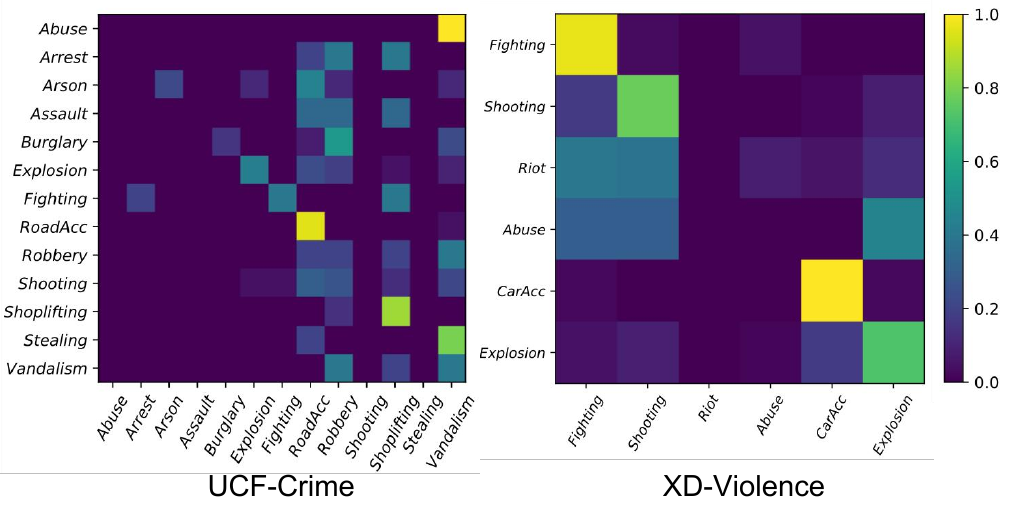}  
\caption{Confusion matrices of anomaly categorization.}\label{mat}
\vspace{-0.5cm}
\end{figure}

We first present qualitative detection results on three datasets in~\cref{vis}, where the top column denotes UCF-Crime, the first three in the bottom column denote XD-Violence, and the rest denotes UBnormal. As we can see, whether base or novel categories, our method produces high anomaly confidence in anomaly regions, even there are multiple discontinuous abnormal regions in a video. Besides, we present confusion matrices of anomaly categorization in~\cref{mat}, it is not hard to see that there are some anomaly categories that our model cannot effectively identify, either base or novel, especially on UCF-Crime, such results indicate OVVAD is a unique and challenging task, especially for the anomaly categorization. Refer to supplementary materials for more ablation studies and qualitative results.

\section{Conclusion}\label{sec5}
In this paper, we present a new model built on top of pre-trained large models for open-vocabulary video anomaly detection task under weak supervision. Owing to the challenging nature of open-vocabulary video anomaly detection, current video anomaly detection approaches face difficulties in working efficiently. To address these unique challenges, we explicitly disentangle open-vocabulary video anomaly detection into the class-agnostic detection and class-specific classification sub-tasks. We then introduce several ad-hoc modules: temporal adapter and semantic knowledge injection modules mainly aim at promoting detection for both base and novel anomalies, novel anomaly synthesis module generates several potential pseudo novel sample to assist the proposed model in perceiving novel anomalies more accurately. Extensive experiments on three public datasets demonstrate the proposed model performs advantageously on open-vocabulary video anomaly detection task. In the future, generating more vivid pseudo anomaly samples in the form of videos with the assistance of AIGC models is yet to be explored.

\section{Acknowledgments}
This work is supported by the National Natural Science Foundation of China (No. 62306240, U23B2013), China Postdoctoral Science Foundation (No. 2023TQ0272), National Key R\&D Program of China (No.2020AAA0106900), and the Fundamental Research Funds for the Central Universities (No. D5000220431).
{
    \small
    \bibliographystyle{ieeenat_fullname}
    \bibliography{main}
}

\end{document}